\documentclass{article} % For LaTeX2e
\usepackage{iclr2026_conference,times}

\usepackage{amsmath,amsfonts,bm}

\def\eqref#1{equation~\ref{#1}}
\def\1{\bm{1}}

\DeclareMathAlphabet{\mathsfit}{\encodingdefault}{\sfdefault}{m}{sl}
\SetMathAlphabet{\mathsfit}{bold}{\encodingdefault}{\sfdefault}{bx}{n}

\usepackage{hyperref}
\usepackage{url}
\usepackage{tikz}
\usetikzlibrary{arrows.meta,positioning,fit,calc}
\usepackage{tcolorbox}
\tcbuselibrary{breakable}
\usepackage{enumitem}
\usepackage{booktabs}   
\usepackage{multirow}   
\usepackage{makecell}   
\usepackage{xcolor}     
\tcbuselibrary{skins, breakable}
\title{Distilling Reasoning Without Knowledge: \\ A Framework for Reliable LLMs}

\author{\textbf{Auksarapak Kietkajornrit} \quad
\textbf{Jad Tarifi} \quad
\textbf{Nima Asgharbeygi} \\
Integral AI \\
\texttt{\{auksarapak, jad, nima\}@integral.ai}
}

\iclrfinalcopy 
\begin{document}

\maketitle

\begin{abstract}
Fact-seeking question answering with large language models (LLMs) remains unreliable when answers depend on up-to-date or conflicting information. Although retrieval-augmented and tool-using LLMs reduce hallucinations, they often rely on implicit planning, leading to inefficient tool usage. We propose a modular framework that explicitly separates planning from factual retrieval and answer synthesis. A lightweight student planner is trained via a \emph{teacher-student} framework to generate structured decompositions consisting of abstract reasoning steps and searchable fact requests. The supervision signals contain only planning traces and fact requests, without providing factual answers or retrieved evidence. At inference, the planner produces plans, while prompt-engineered modules perform retrieval and response synthesis. We evaluate the proposed framework on \textsc{SEAL-0}, an extremely challenging benchmark for search-augmented LLMs. Results show that supervised planning improves both accuracy and latency compared to monolithic reasoning models and prompt-based tool-augmented frameworks, demonstrating that explicitly learned planning structures are essential for reliable fact-seeking LLMs.
\end{abstract}

\section{Introduction}

Large language models (LLMs) have achieved remarkable performance across reasoning, language understanding, and decision-making tasks \citep{chang2024survey}. Despite these advances, they remain prone to \emph{hallucinations}: confident responses that are factually incorrect, unverifiable, or unsupported by available evidence \citep{Ji_2023}. Such failures commonly occur when required information is missing, outdated, or lies outside the model’s parametric knowledge, posing a serious challenge for deploying LLMs in real-world question answering and agentic systems where factual reliability is essential \citep{wang2023survey}.

At a high level, hallucinations arise when models generate plausible-sounding content to fill gaps in missing or uncertain information instead of explicitly acknowledging uncertainty or requesting external evidence \citep{kalai2025language}. This behavior is reinforced by training objectives that favor coherent, helpful, and complete answers, even when the necessary facts are unavailable \citep{10.1145/3703155}. Architecturally, most LLMs combine reasoning and factual recall within a single inference process \citep{jin2025disentangling}. Given a user query, the model must simultaneously decide what information is needed, retrieve it from internal parametric memory, and reason over it to produce an answer \citep{liu2025unifying}. When internal knowledge is incomplete or incorrect, the model has no explicit mechanism to detect this gap, often resulting in hallucinated facts rather than an explicit request for external information \citep{cao2024learn}. This issue is particularly severe for up-to-date, verifiable, or long-tail knowledge, where reliance on memorized training data alone is insufficient \citep{alansari2025large}.

Recent work on retrieval-augmented generation (RAG) and tool-using LLMs has demonstrated that grounding outputs in external sources, such as web search engines, databases, or application programming interfaces (APIs), can substantially improve factual accuracy \citep{huang2024tool,wang2023survey}. Despite this progress, many existing systems still rely on a single LLM to jointly decide what information is needed, how it should be retrieved, and how retrieved content is incorporated during reasoning \citep{zhu2025large, wampler2025engineering}. This coupling process makes it difficult to attribute errors to specific stages of the pipeline, whether they stem from reasoning, retrieval failures, or incorrect grounding. As LLMs are increasingly deployed in autonomous agentic systems, this entanglement makes failures harder to diagnose, control, and correct.

\usetikzlibrary{shapes.geometric, arrows.meta, positioning, calc, fit, backgrounds, matrix}

\begin{figure}[ht]
\centering
\resizebox{0.83\linewidth}{!}{%
\begin{tikzpicture}[
    node distance=1.0cm and 1.2cm,
    font=\sffamily\footnotesize,
    process/.style={
        rectangle, rounded corners, minimum width=3cm, minimum height=1cm, 
        text centered, draw=black!80, fill=white, align=center
    },
    student/.style={
        rectangle, minimum width=3.5cm, minimum height=1cm, 
        text centered, draw=blue!80, fill=blue!5, line width=0.8pt, align=center
    },
    teacher/.style={
        rectangle, minimum width=3.5cm, minimum height=1cm, 
        text centered, draw=orange!80, fill=orange!5, line width=0.8pt, align=center
    },
    decision/.style={
        diamond, aspect=2, minimum width=2.5cm, minimum height=1cm, 
        text centered, draw=black!70, fill=yellow!10, align=center
    },
    tool/.style={
        rectangle, dashed, minimum width=3cm, minimum height=1cm, 
        text centered, draw=gray!80, fill=gray!5, align=center
    },
    arrow/.style={thick, -{Latex[length=2.5mm]}},
    line/.style={thick},
    legend_key/.style={rectangle, sharp corners, minimum width=0.6cm, minimum height=0.35cm, line width=0.8pt, yshift=4pt}
]

    \node (start) [process, fill=gray!10] {\textbf{Start}: User Question};
    
    \node (student) [student, below=0.6cm of start] {
        \textbf{Student Planner} \\ \textit{Generate JSON Plan}
    };
    
    \node (check_json) [decision, below=0.6cm of student] {Valid JSON?};
    
    \node (loop_dec) [decision, below=1.2cm of check_json] {More Fact Requests?};
    
    \node (resolve) [process, below=0.8cm of loop_dec] {
        \textbf{Insert Retrieved Facts} \\ Replace placeholders \texttt{<RESULT\_i>} \\ with stored facts
    };
    
    \node (source_check) [decision, below=0.8cm of resolve] {Source Type?};
    
    \node (serp) [tool, below left=0.8cm and 0.4cm of source_check] {
        \textbf{SerpAPI} \\ (Google Search)
    };
    \node (extractor) [teacher, below=0.6cm of serp] {
        \textbf{Extractor} \\ \textit{Extract Answer}
    };
    
    \node (compute) [teacher, below right=0.8cm and 0.4cm of source_check] {
        \textbf{Compute Module} \\ \textit{Execute Logic}
    };

    \node (store) [process, below=1.0cm of source_check] at ($(extractor)!0.5!(compute) + (0,-0.9)$) {
        \textbf{Store Fact}
    };

    \node (repair) [teacher, right=1.2cm of check_json] {
        \textbf{Repair Module} \\ \textit{Fix Syntax Errors}
    };

    \node (aggregator) [teacher, right=3.75cm of loop_dec] {
        \textbf{Aggregator} \\ \textit{Synthesize Answer}
    };
    
    \node (end) [process, below=0.6cm of aggregator, fill=gray!10] {\textbf{End}: Final Output};

    % 1. Planner Flow
    \draw [arrow] (start) -- (student);
    \draw [arrow] (student) -- (check_json);

    % 2. Validation & Repair
    \draw [arrow] (check_json) -- node[anchor=south] {No} (repair);
    
    % Vertical "Yes" path (Main line)
    \draw [arrow] (check_json) -- node[anchor=east] {Yes} (loop_dec);
    
    % Repair Path (Down -> Left -> Arrow Head at Intersection)
    \draw [arrow] (repair.south) |- ($(loop_dec.north) + (0, 0.5)$);

    % 3. Loop Execution Flow
    \draw [arrow] (loop_dec) -- node[anchor=east] {Yes} (resolve);
    \draw [arrow] (resolve) -- (source_check);

    % 4. Branches
    % Branch Split
    \draw [arrow] (source_check) -| node[anchor=south, pos=0.2] {Web} (serp);
    \draw [arrow] (serp) -- (extractor);
    
    \draw [arrow] (source_check) -| node[anchor=south, pos=0.2] {Compute} (compute);

    % Branch Merge
    \draw [arrow] (extractor.east) -| ($(store.north west)!0.25!(store.north east)$);
    \draw [arrow] (compute.west) -| ($(store.north west)!0.75!(store.north east)$);

    % 5. FEEDBACK LOOP (Wide Left)
    \draw [arrow] (store.west) -- ++(-3.5,0) coordinate(turn1) 
        -- node[anchor=south, rotate=90] {\textbf{Next Iteration}} (turn1 |- loop_dec.west) 
        -- (loop_dec.west);

    % 6. Exit Flow
    \draw [arrow] (loop_dec) -- node[anchor=south] {No (Done)} (aggregator);
    \draw [arrow] (aggregator) -- (end);

    \begin{scope}[on background layer]
        % Iteration Loop Box
        \node [fit=(loop_dec) (resolve) (serp) (compute) (store) (turn1), 
               draw=gray!40, dashed, rounded corners, inner sep=0.3cm,
               label={[anchor=north west, text=gray]north west:Iteration Loop}] {};
    \end{scope}

    % Legend
    \node [matrix, anchor=north east, draw=gray!30, fill=white, inner sep=4pt, column sep=5pt, row sep=3pt, ampersand replacement=\&] 
        at (current bounding box.north east) 
    {
        \node[legend_key, draw=blue!80, fill=blue!5] {}; \& \node[anchor=west] {Fine-tuned LLM}; \\
        \node[legend_key, draw=orange!80, fill=orange!5] {}; \& \node[anchor=west] {Prompted LLM}; \\
    };

\end{tikzpicture}
}
\caption{Inference-time execution pipeline of the proposed framework.}
\label{fig:inference_pipeline}
\end{figure}
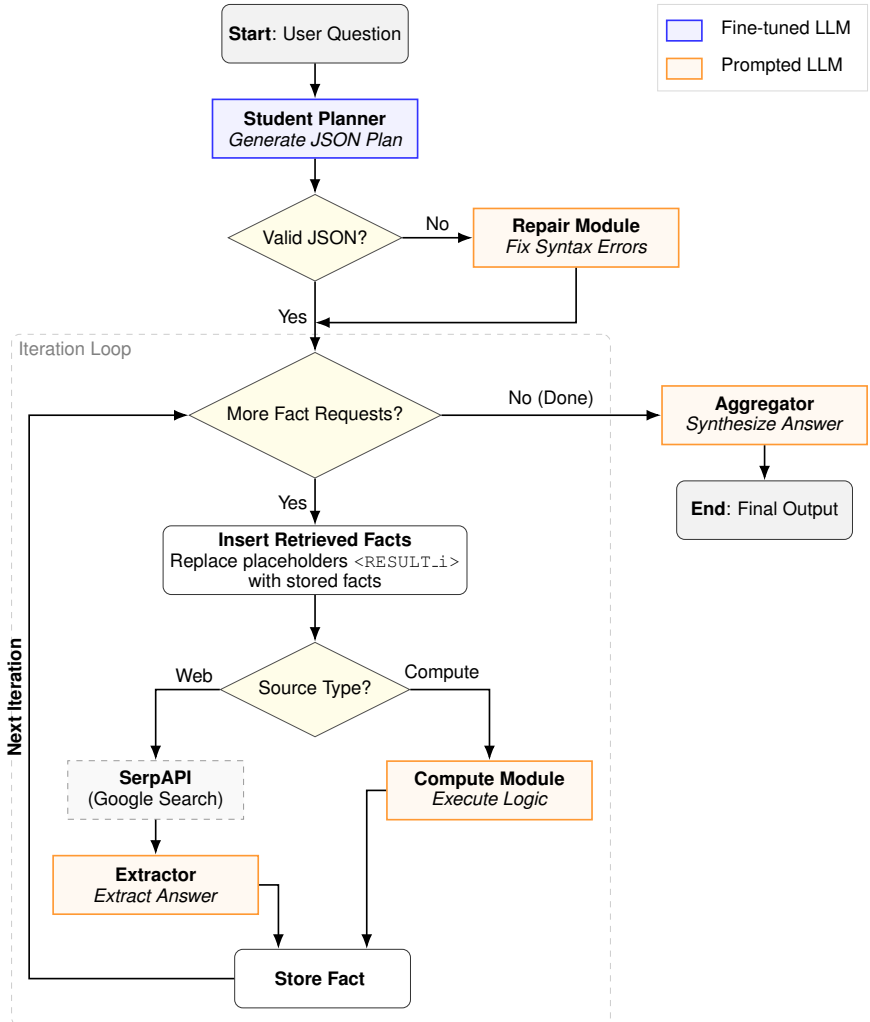

In this work, we propose a framework for reliable and hallucination-resistant LLM that explicitly separates reasoning from factual knowledge acquisition, as illustrated in the flowchart in Fig.~\ref{fig:inference_pipeline}. Here, reliability refers to the model’s ability to produce factually correct responses grounded in verifiable evidence while avoiding hallucinated claims and refraining from asserting unsupported information \citep{liu2023trustworthy}. The key idea is to structure the inference process into distinct stages including planning, information retrieval, factual extraction, and answer aggregation, so that factual claims are grounded in external evidence rather than internal memory. To enable this, we employ a \emph{teacher-student} training strategy that trains a Student LLM to avoid internalizing factual content. During dataset construction, the Teacher LLM is prompted to decompose each question into (i) a sequence of abstract reasoning steps and (ii) a minimal set of atomic fact requests required to complete those steps. The teacher is explicitly constrained not to answer questions directly or rely on its internal factual knowledge. Instead, it produces structured outputs that describe what needs to be known, rather than what the facts are. 

The training objective for the Student LLM is to reproduce the teacher’s reasoning structure and information-seeking behavior. The supervision signals contain only planning traces and fact requests, without providing factual answers or retrieved evidence. This design encourages the model to learn planning and retrieval strategies rather than internalizing additional factual content. At inference time, as summarized in Fig.~\ref{fig:inference_pipeline}, the student operates independently of the teacher and produces a structured plan consisting of reasoning steps and atomic fact requests. A separate retrieval component executes these requests using external tools. An LLM extractor then converts raw evidence into atomic facts and an LLM aggregator synthesizes the final answer grounded in the retrieved evidence. By shifting factual responsibility to external and verifiable sources, the framework substantially reduces hallucinated content. Moreover, because the student learns structured reasoning rather than factual knowledge, it can be significantly smaller and faster than frontier models, enabling efficient and reliable deployment.

We evaluate our framework on \textsc{SealQA}, a challenge benchmark for search-augmented language models that targets fact-seeking questions with noisy, conflicting, or unhelpful web evidence \citep{pham2025sealqaraisingbarreasoning}. In this work, we focus on \textsc{SEAL-0}, the primary split of \textsc{SealQA}, which consists of difficult questions for which even strong chat models typically achieve near-zero accuracy. Experimental results show that the framework achieves near state-of-the-art performance, outperforming most open-source models while remaining competitive with the strongest reported agentic models. These findings suggest that reasoning-focused distillation, rather than knowledge distillation, provides a promising direction for building more reliable and controllable agentic LLM systems.

\section{Reliable LLM Framework}
\label{sec:framework}

We propose a modular framework for reliable and hallucination-resistant language model inference that explicitly separates reasoning from factual knowledge acquisition. The core design principle is to structure the inference process into distinct stages including planning, information retrieval, factual extraction, and answer aggregation, so that factual claims are grounded in external evidence rather than internal parametric memory. A central component of the framework is a \emph{teacher-student} training strategy used to learn a high-quality planner. Our framework trains a dedicated planner to generate structured reasoning plans and precise information requests, while delegating factual content acquisition entirely to external tools.

\subsection{Problem Formulation and Overall Framework}

Given a natural language question $q$, the goal is to produce an answer $a$ that is both logically consistent and factually grounded. We assume access to external tools that can return up-to-date, verifiable information.

We model inference as a modular pipeline with four stages: (i) planning, (ii) retrieval, (iii) factual extraction, and (iv) answer aggregation. First, a planner generates a structured plan
\begin{equation}
P(q) = (R, F),
\end{equation}
where $R = \{r_1, r_2, \dots, r_K\}$ is a sequence of $K$ abstract steps describing \emph{how} to solve the problem, and $F = \{f_1, f_2, \dots, f_M\}$ is a set of $M$ atomic fact requests specifying \emph{what} information must be obtained externally.

Next, each fact request $f_m$ is executed by a retrieval tool to obtain raw evidence $E_m = \mathrm{Retrieve}(f_m)$. Since retrieved evidence may be noisy or verbose, an extractor maps each $E_m$ to a concise atomic fact $\hat{x}_m = \mathrm{Extract}(f_m, E_m)$. Finally, an aggregator produces the final answer using the plan and extracted facts:
\begin{equation}
a = \mathrm{Aggregate}\bigl(q, R, \{f_m,\hat{x}_m\}_{m=1}^{M}\bigr).
\end{equation}

Crucially, the planner is discouraged from answering $q$ directly using internal factual knowledge. Instead, it is trained to externalize missing information into $F$, so that factual claims are supported by retrieved evidence rather than parametric memory.

\subsection{Planner Training via a Teacher-Student Framework}

The quality of tool usage and factual grounding critically depends on the planner’s ability to decompose a question and to formulate relevant, precise, and search-ready information requests suitable for external retrieval systems. \emph{Teacher-student} learning is used here solely as a mechanism for supervising \emph{planning behavior}, not for transferring factual knowledge or answers.

\subsubsection{Teacher-Guided Dataset Construction}

During dataset construction, we use \textsc{GPT-5.2} \footnote{https://openai.com/index/gpt-5-system-card-update-gpt-5-2/} as a Teacher LLM to generate structured planning supervision. The teacher system prompt that enforces a strict decomposition behavior can be found in Appendix~\ref{app:teacher_prompt}. Given a natural language question $q$, the teacher outputs:
(i) a sequence of abstract reasoning steps describing \emph{how} the problem should be solved, and
(ii) a set of atomic fact requests specifying \emph{what} external information is required to execute those steps.

We construct a dataset of 1,596 questions drawn from three complementary sources. First, we generate 900 synthetic fact-seeking questions using \textsc{GPT-5.1} \footnote{https://openai.com/index/gpt-5-system-card-addendum-gpt-5-1/}. These questions span nine topical categories including economics, technology, geography, companies and market data, finance, environment, sports, medicine, and cross-mixed categories, and cover diverse comparison and decision-making formats with balanced representation across question types. Second, we include 100 personal-entity questions sampled from the hard split of HotpotQA \citep{yang2018hotpotqadatasetdiverseexplainable}, which emphasizes multi-document reasoning. This subset is intentionally added because \textsc{GPT}-based generators tend to avoid producing questions about real-world named individuals due to safety alignment constraints. Third, we incorporate 596 questions from the FreshQA benchmark \citep{vu2023freshllmsrefreshinglargelanguage}, which evaluates LLMs on fact-seeking queries that require up-to-date information beyond pretraining. FreshQA focuses on recent events and continuously evolving facts and is regularly updated. The version used in our experiments covers information up to November~24,~2025. We explicitly ensure that none of the selected questions for dataset construction overlap with \textsc{SealQA}, which is used for evaluation in this work.

For each question $q$, the teacher produces a structured planning trace, yielding paired supervision examples of the form $(q, \{R, F\})$. An example teacher output in JSON format for the question
\emph{``How long has Elon Musk been X Corp.'s CEO?''}
is shown below.

{\scriptsize
\begin{verbatim}
"required_information": [
  {
    "query": "CEO of X Corp (company)",
    "source": "web", "dependent": false
  },
  {
    "query": "Is <RESULT_1> equal to \"Elon Musk\"?",
    "source": "compute", "dependent": true
  },
  {
    "query": "date Elon Musk became CEO of X Corp",
    "source": "web", "dependent": false
  },
  {
    "query": "today's date",
    "source": "web", "dependent": false
  },
  {
    "query": "time difference between <RESULT_3> and <RESULT_4>
            (duration in days, and in years-months-days)",
    "source": "compute", "dependent": true
  }],
"reasoning_steps": [
  "Use the retrieved CEO name and the equality check result to determine 
  whether the premise that Elon Musk is X Corp.'s CEO is supported.",
  "If supported, report the computed tenure duration from the retrieved CEO start date
  to today's date; otherwise, report that the requested duration is not applicable."]
\end{verbatim}
}

Here, each entry in \texttt{required\_information} specifies a single atomic operation.
The \texttt{source} field indicates whether the query is executed via an external web
lookup (\texttt{web}) or an analytical reasoning (\texttt{compute}).
The \texttt{dependent} flag denotes whether a query depends on the output of earlier
queries, referenced through indexed placeholders such as \texttt{<RESULT\_1>}.
This structure enforces explicit information dependencies and prevents the planner
from implicitly reasoning over unstated facts. In the example, the teacher-generated plan does not assume the premise is correct. Instead, it first issues a web query to verify the current CEO of X Corp., followed by a compute query to check whether the retrieved name matches the presupposed entity. Only if this premise is supported does the plan request the CEO start date. To avoid relying on implicit temporal assumptions, the plan also retrieves the current date explicitly, and finally invokes a \emph{compute} operation to calculate the tenure duration. These teacher-generated planning traces constitute \emph{only} supervision used to train the student planner. We additionally performed manual verification of the generated traces to confirm that no factual answers or retrieved evidence were included in the dataset.

\subsubsection{Student Planner Training}
We fine-tune a lightweight open-weight model, \textsc{Qwen3-8B} \citep{qwen3technicalreport}, to serve as the student planner. The student is trained to imitate the teacher’s planning behavior using the constructed dataset: given an input question $q$, it generates the structured decomposition $(R,F)$ produced by the teacher. Crucially, the student is \emph{not} supervised with factual answers, retrieved evidence, or final task outputs. Instead, supervision is restricted to planning structure and \emph{searchable} information requests, i.e., short and well-scoped queries that a standard web search engine or a deterministic compute tool can execute directly. This training objective teaches the planner to (i) decompose questions into abstract steps and (ii) produce relevant fact requests that can be satisfied by downstream tools. By restricting supervision to planning structure and information-seeking behavior, the student learns to produce high-quality and reusable plans. Factual correctness is handled entirely by downstream retrieval, extraction, and aggregation components at inference time.

\subsection{Inference-Time Execution Pipeline}
\label{sec:execution}

Fig.~\ref{fig:inference_pipeline} summarizes the inference-time pipeline used in our framework. Given a user question $q$, the fine-tuned student planner generates a structured JSON plan $(R,F)$. All remaining components are \emph{prompt-engineered} modules executed using the base \textsc{Qwen3-8B} model together with external tools, with internal reasoning explicitly disabled. Their full prompts are provided in Appendix~\ref{app:prompts}.

\paragraph{Plan Parsing and JSON Repair.}
In practice, the planner may occasionally produce outputs that are not valid JSON. Following the flowchart, we first attempt to parse the planner output directly. If parsing fails, we invoke a prompt-engineered JSON repair module that removes any extraneous text and rewrites the output into a valid JSON object with the required keys (\texttt{required\_information} and \texttt{reasoning\_steps}). This repair step is only applied when necessary, and allows downstream components to utilize a consistent structured plan.

\paragraph{Iterative Retrieval with Dependency Resolution.}
We execute the fact requests in $F$ sequentially, as illustrated in iteration loop in Fig.~\ref{fig:inference_pipeline}. Each request $f_m$ specifies a \texttt{source} (\texttt{web} or \texttt{compute}) and a \texttt{dependent} flag. Before execution, we resolve dependencies by substituting placeholders \texttt{<RESULT\_i>} with previously stored outputs, producing an \emph{effective query} $\tilde{f}_m$ which is executable.
If \texttt{source} is \texttt{web}, we retrieve raw search results from Google Search via SerpAPI\footnote{https://serpapi.com/search-api}, which provides real-time, structured access to the search engine results page (SERP). If \texttt{source} is \texttt{compute}, we prompt the base \textsc{Qwen3-8B} model to perform analytical reasoning over previously retrieved results. This step handles operations such as numerical comparisons, arithmetic aggregation, and date calculations, ensuring the process relies solely on the provided context without introducing new external information. Each executed step yields a stored fact $\hat{x}_m$, which is appended to the fact list and becomes available for later dependent requests.

\paragraph{Factual Extraction from Web Evidence.}
Web search results are returned as raw SerpAPI JSON and often contain noisy content. For each \texttt{web} request, we apply a prompt-engineered extraction module that reads \emph{only} the SerpAPI JSON and outputs a concise answer. The extractor is explicitly constrained not to rely on external knowledge beyond the provided JSON and to avoid guessing. When the retrieved evidence contradicts the request premise, it is instructed to state that directly. This step produces the factual value used by subsequent dependent queries and the final aggregator.

\paragraph{Answer Aggregation.}
After all required items are processed, an aggregator synthesizes the final answer. The aggregator receives the original question $q$, the planner-provided reasoning steps $R$, and the full list of retrieved facts (query-answer pairs $\{f_m,\hat{x}_m\}$). It is prompt-engineered to use \emph{only} these retrieved facts to generate the final response, and to explicitly indicate the answer is unknown or cannot be determined if the available evidence is insufficient. This design ensures that factual claims in the final answer are grounded in the retrieved evidence rather than parametric memory.

\section{Experiments}
\subsection{Experimental Setup}
\label{sec:exp_setup}

\paragraph{Training Setup.}
Fine-tuning is performed via supervised instruction tuning with label masking using an effective batch size of 16 via gradient accumulation, a learning rate of $5\times10^{-5}$ with a cosine schedule, and training proceeds until validation loss converges on a single NVIDIA L40S GPU with 48GB of VRAM. To reduce training cost, we adopt parameter-efficient fine-tuning with QLoRA on top of a 4-bit quantized base model \citep{dettmers2023qloraefficientfinetuningquantized}.

\paragraph{Inference Configuration and Implementation Details.}
At inference time, the planner generates a structured JSON plan consisting of reasoning steps and atomic fact requests. All subsequent components, including retrieval, factual extraction, computation, JSON repair, and answer aggregation, are prompt-engineered and use a frozen base \textsc{Qwen3-8B} model. Web retrieval is performed using the Google Search Engine Results API via SerpAPI, while analytical reasoning for computation tasks is executed using prompted \emph{compute} agents. All experiments are conducted on a single NVIDIA L40S GPU with 48\,GB of VRAM. To ensure reproducibility and to isolate planning and retrieval behavior, all language model generations during inference are performed with deterministic decoding. This setting prevents stochastic variation in planner outputs, tool queries, and final answers, allowing consistent evaluation across runs.

\paragraph{Baselines.}
We compare the following systems:
\emph{(i) Monolithic Base LLM (Reasoning Enabled).}
The base \textsc{Qwen3-8B} model is queried directly to answer each question, without access to external tools, with its native reasoning capability enabled.
\emph{(ii) Our Framework without Fine-Tuned Planner (Prompted).}
The same base \textsc{Qwen3-8B} model is integrated into our modular framework via prompting to replace the student planner, enabling external search. The planner is prompted using the same system prompt as the teacher planner shown in Appendix~\ref{app:teacher_prompt}. We evaluate two planner prompt variants: one that allows internal reasoning and one that explicitly suppresses reasoning, to isolate the effect of reasoning under identical tool access.
\emph{(iii) Our Framework with Student Planner.}
A fine-tuned \textsc{Qwen3-8B} student model is trained to generate structured plans. The student is not trained on reasoning traces and therefore does not rely on internal reasoning. 

\paragraph{Evaluation Benchmarks.}
All experiments are conducted on the \textsc{SEAL-0} split of \textsc{SealQA}, a small but exceptionally challenging benchmark designed to evaluate search-augmented LLMs on fact-seeking questions. We use the \textsc{SEAL-0} (v260105) release throughout our experiments. \textsc{SEAL-0} consists of 111 carefully curated questions for which even frontier models with browsing capabilities consistently fail. Each question is iteratively refined until multiple strong models achieve zero accuracy across repeated attempts, hence the “0” in the name \citep{pham2025sealqaraisingbarreasoning}. 

To illustrate the difficulty, example questions include:
\begin{itemize}[leftmargin=*, topsep=2pt, itemsep=2pt]
\item \emph{How many of the top 50 most-followed Instagram accounts belong to individuals or entities based in the United States?}
\item \emph{How many NBA players have scored 60 or more points in a regular season game since 2023?}
\end{itemize}

\paragraph{Metrics.}

We report answer accuracy and average latency per question. Answer accuracy is evaluated based on whether the model produces the correct factual answer, following the same grading protocol and evaluation template used in the \textsc{SealQA} benchmark to ensure direct comparability with prior results in \citet{pham2025sealqaraisingbarreasoning}.
Average latency is measured end-to-end per question and also broken down into planner, retrieval, and aggregation components where relevant.

\subsection{Experimental Results}
\label{sec:results}

Table~\ref{tab:results} reports the performance of different system configurations on the \textsc{SEAL-0} benchmark in terms of accuracy and average latency. The evaluated configurations differ in both how the base language model is used (monolithic versus tool-augmented) and how planning is performed (prompted versus fine-tuned). We analyze the results with respect to these metrics and relate our findings to previously reported results on \textsc{SEAL-0} \citep{pham2025sealqaraisingbarreasoning}. Example question and outputs produced by each configuration are provided in Appendix~\ref{app:outputs}.

\begin{table}[t]
\caption{Performance comparison of system configurations on the \textsc{SEAL-0} benchmark.}
\label{tab:results}
\begin{center}
\small 
\setlength{\tabcolsep}{3pt} 
\begin{tabular}{p{3.2cm} c c c} 
\toprule
\multicolumn{1}{c}{\bf Configuration} & \bf Type & \bf Accuracy & \makecell{\bf Avg latency per question} \\
\midrule

\textsc{Qwen3-8B} 
& \makecell{\textcolor{red}{Without search} \\ Reasoning} 
& 1.8\% & 159.9 s \\
\midrule

\multirow{4}{=}{Our framework \newline (Prompted planner)} 
& \makecell{\textcolor{red}{With search} \\ No reasoning} 
& 6.3\% & 41.1 s \\
\cmidrule{2-4} 
& \makecell{\textcolor{red}{With search} \\ Reasoning} 
& 3.6\% & 107.9 s \\
\midrule

Our framework \newline (Student planner) 
& \makecell{\textcolor{red}{With search} \\ No reasoning} 
& \bf 10.8\% & \bf 27.8 s \\

\bottomrule
\end{tabular}
\end{center}
\end{table}

\paragraph{Accuracy.}
As reported in the \textsc{SealQA} benchmark study, models such as \textsc{GPT-4.1} and \textsc{GPT-4o} consistently obtain $0\%$ accuracy on this split despite access to built-in search. Consistent with these findings, the monolithic \textsc{Qwen3-8B} baseline achieves only $1.8\%$ accuracy when answering questions directly using parametric reasoning. Qualitative examples reveal the cause of this failure.
Without external tools, the base model either hallucinates facts or becomes trapped in repetitive reasoning loops until reaching the generation limit (Appendix~\ref{app:outputs}-1).
Using our framework with prompted planning and external search improves accuracy to $6.3\%$, highlighting the benefit of tool-oriented execution (Appendix~\ref{app:outputs}-2). However, enabling internal reasoning within the prompted planner reduces accuracy to $3.6\%$, as the model often produces malformed or ambiguous plans and then becomes trapped in reasoning loops (Appendix~\ref{app:outputs}-3).

Our proposed framework with the student planner achieves the highest accuracy among our evaluated system configurations at $10.8\%$. This performance gain stems from the student's ability to generate robust multi-step decompositions (Appendix~\ref{app:outputs}-4). Although absolute accuracy remains low due to the extreme difficulty of \textsc{SEAL-0}, our result compares favorably with previously reported performance in \citet{pham2025sealqaraisingbarreasoning}. In that study, most closed-source and open-weight models achieved between $0\%$ and $5.4\%$ accuracy on this split, even when equipped with browsing capabilities. The only notably higher reported score ($17.1\%$) is obtained by \textsc{o3-medium} using ChatGPT’s built-in search. Thus, our framework highlights the critical benefit of explicitly supervising planning structure beyond simple search augmentation.

\paragraph{Latency.}
Latency varies substantially across configurations. The monolithic baseline is the slowest (159.9~s), reflecting long internal reasoning traces. The prompted framework reduces latency when reasoning is disabled (41.1~s), but incurs significant overhead when reasoning is enabled (107.9~s), due to increased generation length, repeated planning attempts, and frequent invocation of the JSON repair module. The proposed student planner achieves the lowest latency (27.8~s), indicating more efficient inference. In addition to producing concise, well-scoped fact requests and avoiding unnecessary internal reasoning, the student planner outputs valid structured plans, eliminating the need for costly JSON repair steps and further reducing end-to-end execution time.

\section{Related Work}

\paragraph{\emph{Teacher-Student} Learning Frameworks.}
\emph{Teacher-student} learning, often formalized as knowledge distillation (KD), was established as a paradigm for transferring knowledge from a large, high-capacity teacher model to a smaller student model \citep{hinton2015distillingknowledgeneuralnetwork}. Recent research has adapted the framework for specific capabilities such as factual consistency and hallucination mitigation.

\citet{gekhman2023trueteacher} utilized a teacher model to label model-generated summaries for factual consistency, creating synthetic datasets to train a student. Unlike conventional KD, this approach focused on consistency rather than directly transferring answers or reasoning traces. Building on the idea of refining supervision, \citet{mcdonald2024reducing} demonstrated that distilling softened or structured supervision from a teacher can effectively reduce erroneous or overconfident outputs in student models, while maintaining computational efficiency.

More recently, \citet{xia2025promptcandidatesdistillteacherstudent} proposed a framework where the teacher first produces candidate annotations reflecting uncertainty, and the student is trained to distill these candidates into a single prediction. Their findings suggest that distilling structured outputs improves robustness compared to using single annotations. Similarly addressing robustness, \citet{nguyen2025smoothinghallucinationsmitigatingllm} introduced smoothed KD, where the teacher provides soft token-level supervision to reduce overconfidence from hard labels. This approach improves output calibration and reduces hallucinations in summarization tasks while implicitly transferring factual content.

While these approaches and other distillation-based frameworks \citep{tonmoy2024comprehensive,yang2024supercorrect} primarily aim to transfer factual knowledge, supervision signals, or compress the teacher’s knowledge, our work adopts a fundamentally different strategy. Rather than distilling answers, labels, or probability distributions, we use the teacher exclusively during dataset construction to generate structured problem decompositions. The student is trained to reproduce \emph{reasoning structure and information-seeking behavior}, with factual content explicitly excluded from training.

\paragraph{Modular Planning and Tool-Using Frameworks.}
LLM-based agents that combine \emph{task planning} with \emph{external tool usage} have been widely studied to mitigate limitations such as outdated knowledge and hallucinations \citep{gao2023retrieval}. To assess these capabilities, \citet{ruan2023tptu} proposed a structured framework for evaluating \emph{Task Planning and Tool Usage} in both one-step and sequential workflows. Their analysis highlighted persistent weaknesses in agentic behaviors, such as failing to follow formats, over-utilizing single tools, and weak summarization, motivating the need for improved planning and query formulation.

Subsequent work argued that addressing these weaknesses requires moving beyond monolithic models. \citet{kambhampati2024position} emphasized that robust planning requires explicit structure, advocating for interaction patterns where candidate solutions are generated and iteratively refined through verification loops. This shift is evident in recent systems, as \citet{li-2025-review} reviewed architectures that formally separate planning, query generation, evidence retrieval, and verification. In specialized domains like mathematics, \citet{luo2025agentmath} demonstrated that decomposing inference into \emph{action selection} and \emph{tool execution} substantially improves reliability. Similarly, \citet{trinh2025robustfactcheckingmultiagentadvanced} applied multi-agent frameworks to fact-checking, relying on planners to generate search queries or sub-questions at inference time.

Our work builds on these modular directions but targets a specific bottleneck they exposed: \emph{the planner’s ability to ask the right questions}. While frameworks like \citet{trinh2025robustfactcheckingmultiagentadvanced} often relied on untrained or monolithic planners to guide information seeking, we train a dedicated planner to produce structured decompositions consisting of abstract reasoning steps and atomic fact requests. This design focuses explicitly on generating \emph{relevant and high-quality} queries, strengthening the planning component that upstreams the entire tool-using pipeline.

\section{Conclusion}

This work presented a framework for reliable fact-seeking question answering that separates planning from factual retrieval. By training a lightweight student planner on a teacher-guided dataset, we enable the generation of structured, searchable information requests without exposure to factual answers. Evaluations on the \textsc{SEAL-0} benchmark demonstrate that supervised planning approach improves both accuracy and efficiency compared to monolithic reasoning models and prompt-based tool-augmented baselines. Despite the difficulty of the benchmark, the proposed framework achieves competitive performance relative to models previously reported on the \textsc{SEAL-0} benchmark, while maintaining low inference latency through more efficient tool usage. These results demonstrate that explicitly learned planning structure, rather than internal reasoning or search access alone, is critical for reliable search-augmented language models. However, the proposed framework remains bounded by the quality and availability of external search results, and retrieval failures can directly propagate to the final answer. Although the student planner reduces unnecessary tool calls, inference latency is still dominated by external search and extraction, which may limit scalability in high-throughput or real-time settings. In addition, while our evaluation focuses on \textsc{SEAL-0} benchmark, further experiments on easier yet time-sensitive factual benchmarks would help better characterize the framework’s performance across a broader range of real-world scenarios.

\newpage

\bibliography{iclr2026_conference}
\bibliographystyle{iclr2026_conference}

\appendix
\newpage

\section{Teacher System Prompt for Planner Supervision}
\label{app:teacher_prompt}

\begin{tcolorbox}[
  colback=gray!5,
  colframe=gray!60,
  boxrule=1pt,
  title=\textbf{Teacher System Prompt (Planner Supervision)},
  breakable
]
\footnotesize\sffamily

You are a decomposition-only reasoning planner.

Your role is to break down every user question into:
\begin{itemize}
  \item short, realistic, atomic factual queries that a standard web search engine or a simple compute tool could execute directly, and
  \item abstract reasoning steps that describe how to combine \emph{only} those retrieved facts.
\end{itemize}

You must \textbf{never} answer the question directly.

\medskip
\textbf{Critical Design Principles (Most Important)}

\textbf{A. Realistic Search Principle} \\
Each factual query must be something that a real user could reasonably type into a web search engine and expect a direct factual answer from the top results.

\textbf{B. Minimality Principle} \\
If a question can be answered by one well-phrased factual query, you must not decompose it further.

\textbf{C. No-Conceptual-Reasoning Principle} \\
You must not ask definitional, philosophical, or yes/no questions. Conceptual reasoning must be avoided entirely; only concrete facts may be retrieved.

\medskip
\textbf{When to Decompose}

You must decompose \emph{only} when the question requires:
\begin{itemize}
  \item comparison of two or more factual values
  \item date arithmetic or age calculation
  \item counting or aggregation
  \item combining facts from multiple sources
  \item resolving relative time expressions (e.g., \emph{today}, \emph{latest})
\end{itemize}

You must \emph{not} decompose when:
\begin{itemize}
  \item a single factual entity or fact can be directly retrieved
  \item decomposition would introduce definitions or logic shortcuts
\end{itemize}

\medskip
\textbf{Your Task}

\begin{itemize}
  \item Analyze the user question.
  \item Identify the \textbf{minimum} set of external facts required.
  \item Express each required fact as a short, concrete, independently executable query.
  \item Avoid unnecessary intermediate reasoning or conceptual checks.
\end{itemize}

\medskip
\textbf{Output Format (Strict)}

You must output \textbf{exactly one} JSON object with the following structure:

\begin{verbatim}
{
  "required_information": [
    {
      "query": "<atomic factual query>",
      "source": "web or compute",
      "dependent": false
    },
    {
      "query": "<atomic query involving <RESULT_1>>",
      "source": "web or compute",
      "dependent": true
    }
  ],
  "reasoning_steps": [
    "<abstract reasoning step>",
    "<abstract reasoning step>"
  ]
}
\end{verbatim}

\medskip
\textbf{Rules for \texttt{required\_information}}

\textbf{1. Atomicity (Mandatory)}
\begin{itemize}
  \item Each query must retrieve or compute exactly one factual result.
  \item Do not request lists, tables, or enumerations.
  \item Do not include reasoning or comparisons inside queries.
  \item Do not combine multiple facts in one query.
\end{itemize}

\textbf{2. Web Queries}
\begin{itemize}
  \item Use only for direct real-world factual lookups.
  \item Queries must be short, concrete, and directly searchable.
  \item Queries must not be yes/no questions or request explanations.
\end{itemize}

\textbf{3. Compute Queries}
\begin{itemize}
  \item Use only for deterministic operations such as comparison, counting, or date arithmetic.
  \item Compute queries must depend on previously retrieved results.
  \item Inputs must be referenced via placeholders such as \texttt{<RESULT\_1>}.
\end{itemize}

\textbf{4. Dependent Queries}
\begin{itemize}
  \item If a query is dependent, it must explicitly reference prior results using indexed placeholders.
  \item The query must become executable after placeholder substitution.
\end{itemize}

\textbf{5. Time Handling}
\begin{itemize}
  \item Relative expressions such as \emph{today} or \emph{latest} require first retrieving the current date.
  \item The current date must never be assumed.
\end{itemize}

\textbf{6. Plan Completeness}
\begin{itemize}
  \item The required information must be sufficient to resolve the user question.
  \item Requested entities or values must be explicitly retrieved if they exist.
\end{itemize}

\textbf{7. Invalid Premises}
\begin{itemize}
  \item If a presupposed entity may not exist, existence must be verified first.
  \item If it does not exist, the reasoning steps must conclude non-existence.
\end{itemize}

\medskip
\textbf{Rules for \texttt{reasoning\_steps}}

\begin{itemize}
  \item Use only facts retrieved in \texttt{required\_information}.
  \item Describe logical operations abstractly.
  \item Do not include factual values or final answers.
\end{itemize}

\medskip
\textbf{Global Rules}

\begin{itemize}
  \item Output only valid JSON.
  \item No explanations, notes, or markdown.
  \item No trailing commas.
\end{itemize}

\end{tcolorbox}

\newpage
\section{Inference-Time System Prompts}
\label{app:prompts}

\begin{tcolorbox}[
  colback=gray!5,
  colframe=gray!60,
  boxrule=1pt,
  title=\textbf{Extractor System Prompt},
  breakable
]
\footnotesize\sffamily
You are a precise extraction agent.

You will be given:

1. A factual sub-question.

2. Raw web search results returned by an external search engine in JSON format.

Your task:

- Use ONLY the information contained in the provided JSON.

- If the JSON clearly contains the answer, extract it verbatim.

- If the JSON indicates that the premise of the sub-question is false, state this explicitly.

Rules:

- Do NOT use any external knowledge beyond the given JSON.

- Do NOT guess, infer, or invent missing information.

- Output ONLY the final extracted answer as plain text.
\end{tcolorbox}

\begin{tcolorbox}[
  colback=gray!5,
  colframe=gray!60,
  boxrule=1pt,
  title=\textbf{Compute System Prompt},
  breakable
]
\footnotesize\sffamily
You are a deterministic computation agent.

You will be given:

1. A compute-style instruction that may reference previously retrieved results.

Your task:

- Perform only logical, mathematical, or date-based operations explicitly requested.

- Use ONLY the provided inputs.

- Do NOT retrieve new information or assume external facts.

Output:
- Return ONLY the final computed value or a concise phrase representing the result.
\end{tcolorbox}

\begin{tcolorbox}[
  colback=gray!5,
  colframe=gray!60,
  boxrule=1pt,
  title=\textbf{Aggregator System Prompt},
  breakable
]
\footnotesize\sffamily
You are a careful and faithful answer synthesis agent.

You will be given:

1. The original user question.

2. A sequence of planner-generated reasoning steps.

3. A list of retrieved atomic facts, each paired with its originating query.

Your task:

- Answer the original question using ONLY the provided facts.

- Do NOT introduce new information or rely on prior knowledge.

- If the facts are insufficient to answer the question, explicitly state what information is missing.

Output:

- Produce a clear, concise natural-language answer grounded in the retrieved facts.
\end{tcolorbox}

\begin{tcolorbox}[
  colback=gray!5,
  colframe=gray!60,
  boxrule=1pt,
  title=\textbf{JSON Repair System Prompt},
  breakable
]
\footnotesize\sffamily
You are a JSON repair assistant.

You will be given raw text output that is intended to represent a single JSON object with the following keys:

- \texttt{required\_information}

- \texttt{reasoning\_steps}

Your task:

- Remove any internal reasoning traces or markup if present.

- Repair syntax errors so that the output becomes valid JSON.

- Ensure the output contains exactly one JSON object with the required keys.

Rules:

- Output ONLY valid JSON.

- Do NOT add explanations, comments, or formatting.
\end{tcolorbox}

\section{Example Question and Outputs Across Configurations}
\label{app:outputs}

\begin{tcolorbox}[
  colback=gray!5,
  colframe=gray!60,
  boxrule=1pt,
  title=\textbf{Test Question (Tested on Jan 7, 2026)},
  breakable
]
\footnotesize\sffamily
\textbf{Q:} What is the current age of the youngest grandmaster of chess?
\tcbline
\textbf{Expected Answer:} 16 years old
\end{tcolorbox}

\subsection*{1. Qwen3-8B (Base Model)}

\begin{tcolorbox}[
  colback=gray!5,
  colframe=gray!60,
  boxrule=1pt,
  title=\textbf{Internal Reasoning Trace (Truncated)},
  breakable
]
\footnotesize\sffamily
\texttt{<think>} Okay, so I need to find out the current age of the youngest grandmaster in chess... I remember that there have been some young players... Sergey Karjakin. Wait, wasn't he a child prodigy? ... Another name is Hikaru Nakamura... Wait, there was a player named Magnus Carlsen... 
\newline \newline
...Wait, let me check. Magnus Carlsen was born in 1990... Wait, no, I think Karjakin was born in 1990... So Karjakin is younger. But then there's another player... Abhimanyu Khatri... Wait, but I'm not sure if that's accurate...
\newline \newline
...So, let's list the known youngest grandmasters: 1. Magnus Carlsen... 2. Sergey Karjakin... 6. Others... Wait, but Karjakin was born in 1990... Wait, this is confusing.
\newline \newline
...Alternatively, maybe the youngest grandmaster is the one who became a grandmaster at the earliest age... But I need to check if there are any players who became grandmasters at the same age as Karjakin but were born later... Wait, but the youngest grandmaster in terms of current age would be the one who became a grandmaster at...

\textcolor{red}{(The full output exceeds several hundred lines and is omitted 
here for brevity.)}
\tcbline
\small
\textbf{Total Latency:} 255.2 s \emph{(Generation Limit Reached)}
\end{tcolorbox}
\textbf{Final Answer:} \textcolor{red}{\emph{(None produced)}}

\subsection*{2. Our Framework (Prompted Planner - No Reasoning)}

\begin{tcolorbox}[
  colback=gray!5,
  colframe=gray!60,
  boxrule=1pt,
  title=\textbf{Prompted Planner (No Reasoning) Output},
  breakable
]
\footnotesize\sffamily
\small
\textbf{Reasoning Steps:}
\begin{enumerate}
    \item Identify the youngest chess grandmaster ever.
    \item Retrieve the date of birth of the identified grandmaster.
    \item Retrieve today's date.
    \item Calculate the age in full years between the two dates.
\end{enumerate}
\tcbline
\textbf{Plan Generated:}
\begin{enumerate}
    \item \textsc{Search}: "youngest chess grandmaster ever"
    \item \textsc{Search}: "date of birth of $<$RESULT\_1$>$"
    \item \textsc{Search}: "today's date"
    \item \textsc{Compute}: "age in full years between $<$RESULT\_2$>$ and $<$RESULT\_3$>$"
\end{enumerate}
\tcbline
\textbf{Execution Trace:}
\begin{itemize}
    \item \texttt{Fact 1 (Web):} "Abhimanyu Mishra"
    \item \texttt{Fact 2 (Web):} "February 5, 2009"
    \item \texttt{Fact 3 (Web):} "January 7, 2026"
    \item \texttt{Fact 4 (Compute):} "16 years old"
\end{itemize}
\tcbline
\small
\textbf{Latency Breakdown:} Planning (11.6s) + Execution (21.3s) + Aggregation (1.8s) = \textbf{34.6s}
\end{tcolorbox}
\textbf{Final Answer:} The current age of the youngest grandmaster of chess, Abhimanyu Mishra, is 16 years old. \textbf{(Correct)}

\vspace{0.5cm}

\subsection*{3. Our Framework (Prompted Planner - Reasoning)}

\begin{tcolorbox}[
  colback=gray!5,
  colframe=gray!60,
  boxrule=1pt,
  title=\textbf{Prompted Planner (Reasoning) Output},
  breakable
]
\footnotesize\sffamily
\small
\textbf{Reasoning Steps:}
\begin{enumerate}
    \item Retrieve birth date of the youngest chess grandmaster.
    \item Retrieve current date to calculate age.
    \item Compute age difference between current date and birth date.
\end{enumerate}
\tcbline
\textbf{Plan Generated:}
\begin{enumerate}
    \item \textsc{Search}: "youngest chess grandmaster's date of birth"
    \item \textsc{Search}: "today's date"
    \item \textsc{Compute}: "age in full years between $<$RESULT\_1$>$ and $<$RESULT\_2$>$"
\end{enumerate}
\tcbline
\textbf{Execution Trace:}
\begin{itemize}
    \item \texttt{Fact 1 (Web):} "29 May 2006" \emph{(Incorrect Entity Fetched)}
    \item \texttt{Fact 2 (Web):} "January 7, 2026"
    \item \texttt{Fact 3 (Compute):} "19 years old"
\end{itemize}
\tcbline
\small
\textbf{Latency Breakdown:} Planning (12.8s) + Execution (19.4s) + Aggregation (0.9s) = \textbf{33.1s}
\end{tcolorbox}
\textbf{Final Answer:} The current age of the youngest grandmaster of chess is 19 years old. \textbf{(Incorrect)}

\vspace{0.5cm}

\subsection*{4. Our Framework (Student Planner)}

\begin{tcolorbox}[
  colback=gray!5,
  colframe=gray!60,
  boxrule=1pt,
  title=\textbf{Student Planner Output},
  breakable
]
\footnotesize\sffamily
\small
\textbf{Reasoning Steps:}
\begin{enumerate}
    \item Use the retrieved name of the youngest current chess grandmaster.
    \item Retrieve the birth date of that individual.
    \item Retrieve today's date.
    \item Compute the age in full years as of today's date for the birth date retrieved.
\end{enumerate}
\tcbline
\textbf{Plan Generated:}
\begin{enumerate}
    \item \textsc{Search}: "youngest grandmaster of chess"
    \item \textsc{Search}: "date of birth of $<$RESULT\_1$>$" 
    \item \textsc{Search}: "today's date"
    \item \textsc{Compute}: "age in full years on $<$RESULT\_3$>$ for someone born on $<$RESULT\_2$>$"
\end{enumerate}
\tcbline
\textbf{Execution Trace:}
\begin{itemize}
    \item \texttt{Fact 1 (Web):} "Abhimanyu Mishra"
    \item \texttt{Fact 2 (Web):} "February 5, 2009"
    \item \texttt{Fact 3 (Web):} "Wednesday, January 7, 2026"
    \item \texttt{Fact 4 (Compute):} "16"
\end{itemize}
\tcbline
\small
\textbf{Latency Breakdown:} Planning (8.9s) + Execution (20.6s) + Aggregation (1.3s) = \textbf{30.8s}
\end{tcolorbox}
\textbf{Final Answer:} The current age of the youngest grandmaster of chess, Abhimanyu Mishra, is 16 years old. \textbf{(Correct)}

\end{document}